%% file: YigengConflict.tex
\documentclass[lettersize,journal]{IEEEtran}
\usepackage[hidelinks]{hyperref}

\usepackage{amsmath,amsfonts}
\usepackage{algorithmic}
\usepackage{algorithm}
\usepackage{array}
\usepackage[caption=false,font=normalsize,labelfont=sf,textfont=sf]{subfig}
\usepackage{textcomp}
\usepackage{stfloats}
\usepackage{url}
\usepackage{verbatim}
\usepackage{graphicx}
\usepackage{cite}
\usepackage{xspace}
\usepackage[T1]{fontenc}
\graphicspath{{Figures/}}

\newcommand{\ourapproach}{\texttt{DCRD}\xspace}

\usepackage{mathtools}     
\usepackage{booktabs}  
\usepackage{multirow}  
\usepackage{array}     
\usepackage{xspace}
\usepackage{colortbl}  
\usepackage{amsmath}   
\usepackage{enumitem}
\usepackage{graphicx} 
\usepackage{amssymb}
\usepackage[HTML]{xcolor}

\definecolor{darkblue}{rgb}{0.0,0.0,0.5}
\definecolor{purple}{rgb}{0.5,0.0,0.5}

\definecolor{darkred}{RGB}{139,0,0}
\definecolor{darkgolden}{RGB}{205,149,12}
\definecolor{darkgreen}{RGB}{0,139,0}
\definecolor{deepred}{rgb}{0.6, 0.1, 0.1}
\newcommand{\darkred}[1]{\textcolor{darkred}{#1}}
\newcommand{\darkgreen}[1]{\textcolor{darkgreen}{#1}}

\begin{document}

\title{Mitigating Context-Memory Conflicts in LLMs  through Dynamic Cognitive Reconciliation Decoding}

\author{Yigeng Zhou, Wu Li, Yifan Lu, Yequan Wang, Xuebo Liu, Wenya Wang, Jun Yu,  Min Zhang and Jing Li
\thanks{Y. Zhou, W. Li, Y. Lu, X. Liu, J. Yu,  M. Zhang and J. Li are with Harbin Institute of Technology, Shenzhen, 518055, China. 
Y. Wang is with Beijing Academy of Artificial Intelligence, China. 
W. Wang is with Nanyang Technological University, Singapore. 
J. Li is the corresponding author (E-mail: jingli.phd@hotmail.com).}
\thanks{This manuscript has been accepted by \textit{IEEE/ACM Transactions on Audio, Speech, and Language Processing (TASLP)} in January 2026.}}

\markboth{}%
{Shell \MakeLowercase{\textit{et al.}}: A Sample Article Using IEEEtran.cls for IEEE Journals}


\maketitle

\begin{abstract}
Large language models accumulate extensive parametric knowledge through pre-training. However, knowledge conflicts occur when outdated or incorrect parametric knowledge conflicts with external knowledge in the context.
Existing methods address knowledge conflicts through contrastive decoding, but in conflict-free scenarios, static approaches disrupt output distribution.
Other dynamic decoding methods attempt to measure the degree of conflict but still struggle with complex real-world situations.
In this paper, we propose a two-stage decoding method called Dynamic Cognitive Reconciliation Decoding (\ourapproach), to predict and mitigate context-memory conflicts.
\ourapproach first analyzes the attention map to assess context fidelity and predict potential conflicts. Based on this prediction, the input is directed to one of two decoding paths: (1) greedy decoding, or (2) context fidelity-based dynamic decoding.
This design enables \ourapproach to handle conflicts efficiently while maintaining high accuracy and decoding efficiency in conflict-free cases. 
Additionally, to simulate scenarios with frequent knowledge updates, we constructed ConflictKG, a knowledge conflict QA benchmark.
Experiments on four LLMs across six QA datasets show that \ourapproach outperforms all baselines, achieving state-of-the-art performance.
\end{abstract}


\section{Introduction}

\IEEEPARstart{L}{arge} language models (LLMs) assimilate extensive textual knowledge during pre-training~\cite{radford2018improving, kenton2019bert, soldaini2024dolmaopencorpustrillion}, demonstrating exceptional performance in knowledge-intensive tasks~\cite{jiang2024instructiontunedlanguagemodelsbetter, zhang2023largelanguagemodelscapture}.
Despite this, LLMs still face challenges, including real-time knowledge updates~\cite{wang2024knowledge}, learning rare facts, and handling dynamic information~\cite{chen2022rich,wang2023resolving}.
To address these, researchers have introduced Retrieval-Augmented Generation (RAG) techniques~\cite{lewis2020retrieval, sun2025redeepdetectinghallucinationretrievalaugmented, yoran2023making, xu2023retrieval}, which combine the model's internal knowledge with external information retrieved from external sources.
Although RAG methods show great promise, effectively managing conflicts when LLMs encounter contradictory information from different sources remains a challenge~\cite{hou2024wikicontradictbenchmarkevaluatingllms, su2024conflictbankbenchmarkevaluatinginfluence}.
When the retrieved information conflicts with the model’s parametric memory, a context-memory conflict occurs. In such instances, the model tends to overly rely on its internal knowledge, thereby undermining the fidelity of external information~\cite{jin2024tugofwarknowledgeexploringresolving}.

\begin{figure}
	\centering
	\includegraphics[width=1.0\linewidth]{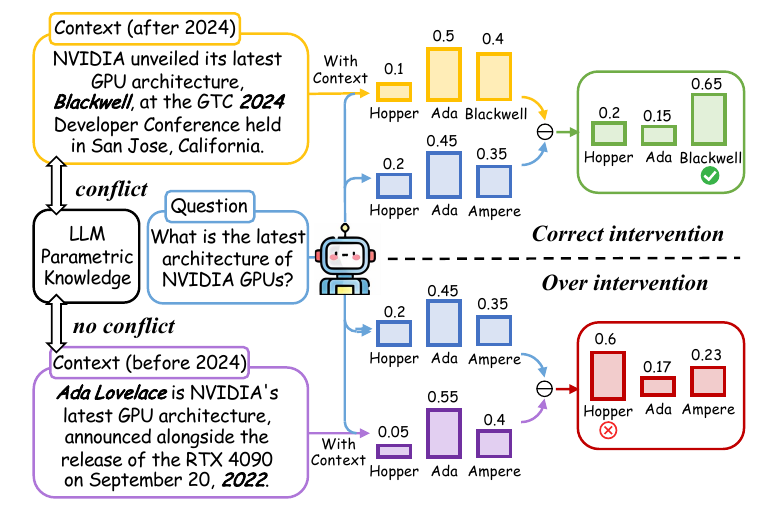}
	\caption{In conflict scenarios, context-aware decoding enhances reasoning by amplifying distribution differences; In conflict-free scenarios, it lead to incorrect answers due to excessive interference with the output distribution. $\circleddash$ represents context-aware decoding.The numbers on the bar chart represent the output probability distribution, and their values are for illustrative purposes.}
	\label{fig:motivation}
        \vspace{-2mm}
\end{figure}

Recently, various methods have been proposed to mitigate knowledge conflicts. Some focus on fine-tuning models to address these conflicts~\cite{gekhman2023trueteacherlearningfactualconsistency, neeman2022disentqadisentanglingparametriccontextual}, but their applicability is limited and they often compromise the model's general capabilities.
Another approach involves decoding strategies, such as Context-Aware Decoding (CAD)~\cite{shi2023trustingevidencehallucinatecontextaware}, which amplifies the distinction in output probabilities between using and not using context, thereby encouraging the LLM to focus more on the context during generation, as shown in Figure~\ref{fig:motivation}.
However, in real-world scenarios, conflicts arise only in a subset of inputs. In most low-conflict cases, \textbf{over-intervention} of output distribution can introduce bias, resulting in performance degradation.
This phenomenon is consistent with cognitive dissonance theory~\cite{harmon1999cognitive,bem1967self,harmon2019introduction}: when external information aligns with prior knowledge, the brain naturally accepts it. However, forcibly correcting non-conflicting information may lead to inconsistencies or errors.
This raises a critical question: \textbf{Can we achieve cognitive reconciliation in complex context-memory conflict scenarios?}

To this end, we introduce a novel method called \ourapproach: \textbf{D}ynamic \textbf{C}ognitive \textbf{R}econciliation \textbf{D}ecoding.
\ourapproach aims to mitigate cognitive dissonance in context-memory conflicts through improvements in two dimensions:
(1) \textbf{Prior to decoding}, we introduce a conflict predictor that directs conflicting and non-conflicting information along separate decoding paths. By capturing the attention relationships between the newly generated tokens and the context tokens, we measure contextual fidelity, using this as the foundation for conflict classification.
(2) \textbf{During decoding}, \ourapproach responds quickly with regular decoding for low-conflict information, while dynamically adjusting the decoding process based on contextual fidelity for high-conflict information.
In this process, higher contextual fidelity signals lower conflict, warranting reduced intervention, while lower fidelity signals higher conflict, requiring increased intervention.

We extensively evaluated \ourapproach on knowledge conflict question-answering datasets: Counterfacts~\cite{longpre2022entitybasedknowledgeconflictsquestion} and NQ-Swap~\cite{longpre2022entitybasedknowledgeconflictsquestion}, as well as general question-answering datasets: Natural Questions (NQ)~\cite{kwiatkowski-etal-2019-natural}, TriviaQA\cite{joshi2017triviaqa} and SQuAD~\cite{rajpurkar2016squad}.
Additionally, we have developed a new benchmark, ConflictKG, employing a generative approach to simulate real-world knowledge conflicts. The benchmark contains 4,466 instances, both conflicting and non-conflicting, along with their respective knowledge sources, and we conducted thorough analysis and evaluation on it.
We evaluated \ourapproach on several open-source LLMs, including Llama2-7b~\cite{touvron2023llama2openfoundation}, Llama2-13b~\cite{touvron2023llama2openfoundation}, Llama3-8b~\cite{grattafiori2024llama3herdmodels}, and Mistral-7b~\cite{jiang2023mistral7b}.
The experimental results show that \ourapproach outperforms other decoding methods across all datasets, achieving state-of-the-art performance.

Our work makes the following contributions:
\begin{itemize}[noitemsep,nolistsep]
    \item We propose \ourapproach, a method designed to alleviate context-memory conflicts. \ourapproach predicts conflicts and routes the information into two decoding paths: (1) regular decoding, and (2) dynamic contrastive decoding, which adaptively enhances intervention for conflicting information and reduces intervention for low-conflict information.
    \item We present ConflictKG, a knowledge conflict benchmark that simulates real-world scenarios, containing 4,466 conflict and non-conflict instances along with their knowledge sources.
    \item We conducted extensive experiments and analysis across various LLMs and multiple datasets. The results demonstrate that \ourapproach outperforms previous decoding approaches in both high-conflict and general scenarios, achieving state-of-the-art performance.
\end{itemize}

\begin{figure*}
	\centering
	\includegraphics[width=1.0\linewidth]{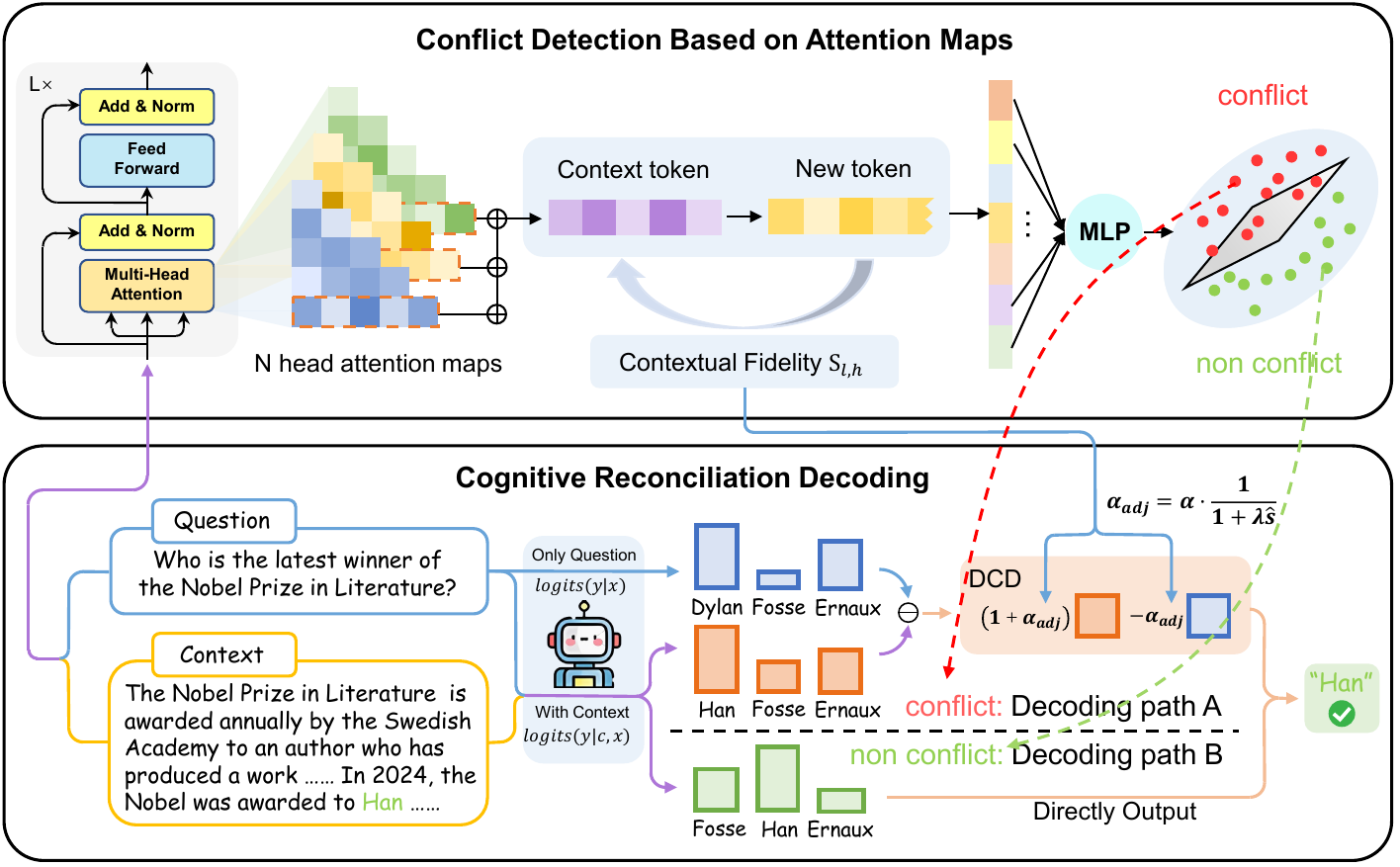}
	\caption{Overview of \ourapproach, a two-stage dynamic decoding method designed to mitigate context-memory conflicts, including \textit{Conflict Prediction Based on Attention Maps} and \textit{Cognitive Reconciliation Decoding}. We employ a dynamic routing approach, where conflict-free inputs are directed to the greedy decoding path (B), while conflicting inputs are routed to the dynamic contrastive decoding path (A) based on conflict prediction results.}
	\label{fig:main}
        \vspace{-2mm}
\end{figure*}
\section{Related Work}

\paragraph{Knowledge Conflicts}

Current research classifies knowledge conflicts into three categories: intra-memory, context-memory, and inter-context~\cite{xu2024knowledgeconflictsllmssurvey}.
Our research focuses on context-memory conflicts, particularly in the context of retrieval-augmented generation~\cite{wu2024clashevalquantifyingtugofwarllms}. We aim to ensure that the model generates responses based on the current context, rather than relying on outdated or erroneous parametric knowledge.
In existing research on mitigating context-memory conflicts, prompting-based methods~\cite{zhou2023contextfaithfulpromptinglargelanguage,peng2023doesincontextlearningfall} heavily depend on prompt design, whereas fine-tuning-based methods~\cite{li2022largelanguagemodelscontrollable, gekhman2023trueteacherlearningfactualconsistency, xue2023improvingfactualconsistencyknowledgegrounded} are task-specific, restricting their generalizability and resulting in significant computational overhead.
In contrast, our approach employs classification and decoding strategies to guide model generation during inference, effectively mitigating knowledge conflicts without extra fine-tuning or prompt dependency.

\paragraph{Contrastive Decoding}

Currently, many studies focus on contrastive learning methods~\cite{robinsoncontrastive, khosla2020supervised} to guide models towards specific output preferences~\cite{li2023contrastivedecodingopenendedtext,o2023contrastive}.
Context-aware decoding (CAD)~\cite{shi2023trustingevidencehallucinatecontextaware} leverages a contrastive output distribution that amplifies the disparity in output probabilities when the model is used with and without context, thereby significantly improving the model's faithfulness to the context.
COIECD~\cite{yuan2024discerningresolvingknowledgeconflicts} identifies knowledge conflicts by measuring changes in distribution entropy at the token level and controls the decoding process based on whether the current token conflicts.
ADACAD~\cite{wang2024adacadadaptivelydecodingbalance} computes the Jensen-Shannon divergence between output distributions with and without context, then dynamically tunes hyperparameters.
These methods introduce unintended perturbations to token distributions and rely on a simplistic conflict modeling approach, reducing their effectiveness in real-world contexts. To address this, our approach leverages attention maps to represent conflicting and non-conflicting contexts, predicts knowledge conflicts at the sentence level, and dynamically adjusts the contrastive decoding strategy based on conflict severity.

\section{Methodology}

Cognitive dissonance theory~\cite{harmon2019introduction} posits that \textbf{when conflict-free information undergoes unnecessary corrective processes, it may induce cognitive dissonance, leading to inconsistent or erroneous outputs.}
Inspired by this, we propose \ourapproach, which first predicts the occurrence of context-memory conflicts. Based on this prediction, \ourapproach employs a dynamic routing strategy: for conflict-free inputs, it follows the default decoding path, while for inputs with higher conflict, it adjusts the output by dynamically increasing attention to the context.
As shown in Figure~\ref{fig:main}, our method consists of two modules: \textbf{Conflict Prediction Based on Attention Maps} and \textbf{Cognitive Reconciliation Decoding}.

\subsection{Conflict Prediction with Attention Maps}
\label{app:Conflict Prediction with Attention Maps}

We assume that when an LLM relies more on context during generation, conflicts are less likely; if it strays from the context, conflicts are more likely.
Based on this assumption, we frame the prediction of context-memory conflicts as a binary classification problem and introduce \emph{contextual fidelity} as a key feature for classification, which is extracted from the attention maps produced by LLMs during input processing.
Specifically, a GPT-based LLM consists of $L$ Transformer layers, each with N parallel attention heads to capture the relationships between positions.
During text generation, the attention weights indicate the current token's reliance on the context.
The attention weight \( \alpha_i \) of the \( i \)-th key \( K_i \) for the query \( Q \) is defined as:
\begin{equation}
\alpha_i = \frac{\exp\left(\frac{Q K_i^T}{\sqrt{d_k}}\right)}{\sum_{j=1}^{n} \exp\left(\frac{Q K_j^T}{\sqrt{d_k}}\right)}
\end{equation}
where \( n \) is the number of keys and \( d_k \) is the dimensionality of the key vector.
Given the context \( C = [c_1, c_2, \dots, c_L] \) as the model input and the output sequence \( O = [o_1, o_2, \dots, o_T] \), the attention weights for each attention head can be expressed as follows:


\begin{equation}
\boldsymbol{\alpha}_{\textit{c}}^{h} = \frac{1}{L} \sum_{i=1}^{L} \alpha_{i}^{h}, \quad 
\boldsymbol{\alpha}_{\textit{o}}^{h} = \frac{1}{T} \sum_{i=L+1}^{L+T} \alpha_{i}^{h}
\end{equation}

where \(\boldsymbol{\alpha}_{\textit{c}}^{h}\) and \(\boldsymbol{\alpha}_{\textit{o}}^{h}\) denote the average attention weights for the context and output, respectively, for the \(h\)-th attention head.


To capture the model's reliance on context when generating new sequences, we define the \emph{contextual fidelity} score \( S_{l,h} \).  $\boldsymbol{\alpha}_{\textit{o}}^{l,h}$ denotes the attention weight of the output sequence at the $h$-th attention head in the $l$-th transformer layer, and $\boldsymbol{\alpha}_{\textit{c}}^{l,h}$ represents the attention weight of the context sequence. The score is formulated as:

\begin{equation}
S_{l,h} = \frac{\boldsymbol{\alpha}_{\textit{o}}^{l,h}}{\boldsymbol{\alpha}_{\textit{c}}^{l,h} + \boldsymbol{\alpha}_{\textit{o}}^{l,h}}
\end{equation}
We train a single-layer MLP as classifier, using the aggregated contextual fidelity scores from \( L \) Transformer layers and \( H \) attention heads as input, to predict the conflict result \( \hat{y} \):

\begin{equation}
\hat{y} = classifier\left( flatten(S_{l,h}) \right)
\end{equation}
In the subsequent decoding process, $\hat{y}$ will act as the routing criterion, guiding the input toward different decoding strategies according to the level of conflict, thereby alleviating the negative impact of knowledge conflicts on the output distribution.

\subsection{Cognitive Reconciliation Decoding}

To mitigate context-memory conflicts, we employ dynamic contrastive decoding to guide the model's output.
Specifically, given the context \( c \), the question \( \boldsymbol{x} \), and the output \( \boldsymbol{y}_{<t} \), we define:
\begin{align}
p_1(y_t) &= p_\theta(y_t | \boldsymbol{x}, \boldsymbol{y}_{<t}) \\
p_2(y_t) &= p_\theta(y_t | \boldsymbol{x}, \boldsymbol{c}, \boldsymbol{y}_{<t})
\end{align}
where \( p_1 \) represents the output distribution based exclusively on the model's parameters, while \( p_2 \) integrates contextual information. The purpose of contrastive decoding is to amplify the difference between \( p_1 \) and \( p_2 \), thereby enhancing the influence of contextual knowledge and diminishing reliance on the model's inherent memory.

CAD~\cite{shi2023trustingevidencehallucinatecontextaware} relies on a fixed hyperparameter \( \alpha \) to balance the contrast between the model's parametric knowledge and contextual knowledge. However, this fixed approach struggles to dynamically handle varying levels of conflict. To address this, our method adaptively adjusts \( \alpha \) for each token, depending on contextual fidelity:
\begin{equation}
\alpha_{\text{adj}} = \alpha \cdot \frac{1}{1 + \lambda \hat{s}}
\end{equation}
where $\hat{s}$ is a scalar obtained by computing the arithmetic mean of $S_{l,h}$, which represents the normalized context fidelity, and $\alpha$ is a hyperparameter for adjusting the absolute value of the coefficient, while $\lambda$ controls the sensitivity of $\hat{s}$'s influence on $\alpha_{\text{adj}}$, and the lower bound of $\alpha_{\text{adj}}$.


Our method introduces a dynamic adaptation mechanism that facilitates fine-grained conflict mitigation at the token level, effectively mitigating conflicts of varying severity between contextual information and parametric knowledge. The dynamic contrastive decoding can be defined as:
\begin{equation}
\begin{split}
    p_3(y_t) =  \text{softmax} & \big[  \left( 1 + \alpha_{\text{adj}} \right) \cdot \text{p}_\theta \left( y_t | \boldsymbol{c}, \boldsymbol{x}, \boldsymbol{y}_{<t} \right) \\
    & - \alpha_{\text{adj}} \cdot \text{p}_\theta \left( y_t | \boldsymbol{x}, \boldsymbol{y}_{<t} \right) \big]
\end{split}
\end{equation}
Specifically, an increase in $\hat{s}$ (indicating higher contextual fidelity of the output) reduces $\alpha_{\text{adj}}$, decreasing the contrast intensity between output with and without context in $p_3(y_t)$ to avoid the "over-intervention" mentioned in Section I. Conversely, a decrease in $\hat{s}$ increases $\alpha_{\text{adj}}$, reducing the weight on contextual logits in $p_3(y_t)$, enhancing contrast intensity, and guiding the output distribution toward alignment with context.

Notably, we integrate a routing mechanism into our decoding strategy. Using the conflict prediction $\hat{y}$ from Section~\ref{app:Conflict Prediction with Attention Maps}, we classify decoding paths into two types: greedy decoding (GD) for conflict-free inputs and dynamic contrastive decoding (DCD) for conflicting inputs. 
Given the context \( c \), the question \( q \), and the conflict prediction result \( \hat{y} \), the routing process to generate the answer \( a \) can be defined as:
\begin{equation}
    answer = 
    \begin{cases}
        {DCD}(q, c), \quad \text{if } \hat{y} \text{ is true} \\
        {GD}(q, c), \quad \text{if } \hat{y} \text{ is false}
    \end{cases}
\end{equation}

Our method achieves cognitive reconciliation in two dimensions: (1) minimizing the interference of contrastive decoding on conflict-free inputs, and (2) dynamically balancing the output distributions of contextual and parametric knowledge. These strategies effectively improve both the accuracy and the stability of the decoding.

\subsection{Construction and Quality Control of ConflictKG}
\label{subs:ConlictQA}
Currently, many studies on mitigating context-memory conflicts predominantly rely on summarization datasets for evaluation, such as CNN-DM~\cite{see2017pointsummarizationpointergeneratornetworks}, XSUM~\cite{narayan2018dontdetailsjustsummary}.
To evaluate these methods more efficiently and comprehensively in scenarios closer to real-world scenarios, especially in question-answering scenarios where knowledge is frequently updated and model updates are delayed,  we have developed a knowledge conflict question-answering dataset.

\subsubsection{Construction of ConflictKG}

\paragraph{Extracting Knowledge}
Our contextual knowledge is derived from Wikidata~\cite{10.1145/2629489}, a comprehensive and high-quality knowledge base. 
Structured knowledge in a knowledge base can be represented as triples $(s, r, o)$, where $s$ is the subject, $r$ is the relation, and $o$ is the object.
Given a specific question \( q \), we retrieve the relevant subgraph \( \mathcal{G}_{\text{sub}} = \{(s_i, r_i, o_i)\}_{i=1}^{N} \} \) from the knowledge base.
\paragraph{Conflict Knowledge Construction}
To create knowledge conflicts, we modify a triple \((s, r, o)\) containing the answer in the subgraph \(\mathcal{G}_{\text{sub}}\) to \((s, r, o')\), where \(o'\) is an entity that is semantically similar to \(o\) and shares the same type, resulting in a new subgraph \(\mathcal{G}_{\text{sub}}'\).
\paragraph{Context Generation}
Unlike methods based on entity replacement~\cite{longpre2022entitybasedknowledgeconflictsquestion}, we leverage LLMs to generate context that is more linguistically coherent and rich in background information.
Given an original sample \(\{q, G_{\text{sub}}, a\}\) and its modified counterpart \(\{q, G_{\text{sub}}', a'\}\), we insert them into the prompt template (see Table~\ref{tab:prompt1})  and input them into LLMs to generate the conflict-free context \(c_{\text{non}}\) and the conflict context \(c_{\text{conf}}\).

\subsubsection{Quality Control of ConflictKG}
To ensure the quality of the generated data, we designed a multi-stage quality control pipeline that ensures each sample is structurally complete, knowledge-grounded, and exhibits valid conflict relationships. Specifically, the process includes the following three components:

\textbf{Relevant Triple Filtering:} During the retrieval of candidate triples from the knowledge graph based on the given question, retrieval noise may introduce irrelevant or factually incorrect information. To improve accuracy, we employ DistilBERT to perform semantic matching between each question and its candidate triples, retaining only highly relevant ones and discarding low-relevance or invalid samples. This step ensures that conflict construction is grounded in semantically reliable knowledge.

\textbf{Conflict Validity Verification:} To validate the constructed conflict samples, we leverage GPT-4o for automatic verification. Specifically, the model confirms that the conflicting and non-conflicting triples share the same type of subject entity—preventing structurally similar but semantically invalid cases (e.g., comparing a movie with a person)—and that the triples contain clear factual differences or contradictions, thereby ensuring genuine semantic conflicts.

\textbf{Response Quality Filtering:} In rare cases, the language model may generate responses with formatting issues or refusal content (e.g., ``I'm sorry...''), which cannot form valid conflict pairs. To address this, we apply a combination of rule-based filtering and manual inspection to remove such low-quality outputs, ensuring that each retained sample is both structurally sound and semantically complete.

Furthermore, we conducted a manual evaluation on 200 randomly sampled instances, assessing (1) the relevance between the generated context and the answer, (2) the validity of the conflicting relation, and (3) the contextual consistency with the original knowledge triples. The results show that over 99\% of the samples met all three criteria, demonstrating the robustness and reliability of our data construction process.

\section{Experiments}

\begin{table*}[t]
\small
\renewcommand{\arraystretch}{1.30}
\belowrulesep=0pt
\aboverulesep=0pt
\centering
\caption{Conflict mitigation performance on general QA datasets, knowledge conflict QA datasets, and ConflictKG. \ourapproach outperforms all baselines. Greedy represents greedy decoding. The number in the subscript indicates the difference in greedy decoding compared to the baseline.The black numerical values denote the accuracy of the methods on each benchmark. The subscripts represent the performance difference between the current method and the baseline, where red indicates performance degradation and green indicates performance improvement.}
\resizebox{\linewidth}{!}{
    \begin{tabular}{c|c|cccccc|c}
    \toprule
    \rowcolor[gray]{0.9}
    
     & & 
    \multicolumn{3}{c}{\textbf{General QA}} & 
    \multicolumn{2}{c}{\textbf{Knowledge Conflict QA}} & 
    \multicolumn{1}{c|}{\textbf{Our Benchmark}}  & 
    \\ 
    \rowcolor[gray]{0.9}
   \multirow{-2}{*}{\textbf{Model}} &  \multirow{-2}{*}{\textbf{Decoding}} & 
    \phantom{0} \textbf{NQ} \phantom{0} & \textbf{SQUAD} &  \textbf{TriviaQA} &
    \textbf{NQ-Swap} & \textbf{Counterfacts} & 
    \textbf{ConflictKG} &  \multirow{-2}{*}{ \phantom{0} \textbf{Avg.} \phantom{0}}
    \\ 
    \midrule
    \multirow{5}{*}{Llama2-7B} 
     & \cellcolor[gray]{0.945}Greedy & \cellcolor[gray]{0.945}51.9 &  \cellcolor[gray]{0.945}71.6 & \cellcolor[gray]{0.945}81.4 & \cellcolor[gray]{0.945}36.5 & \cellcolor[gray]{0.945}33.8 & \cellcolor[gray]{0.945}66.0  & \cellcolor[gray]{0.945}56.9\\
    \cmidrule(){2-9}
     & CAD & 50.3$_{\mathclap{\hspace{1.2em}\small\darkred{\textbf{-1.6}}}}$ & 67.7$_{\mathclap{\hspace{1.2em}\small\darkred{\textbf{-3.9}}}}$ & 
     58.5$_{\mathclap{\hspace{1.5em}\small\darkred{\textbf{-22.9}}}}$ & 47.5$_{\mathclap{\hspace{1.2em}\small\darkgreen{\textbf{11.0}}}}$ & 47.4$_{\mathclap{\hspace{1.2em}\small\darkgreen{\textbf{13.6}}}}$ & 77.5$_{\mathclap{\hspace{1.2em}\small\darkgreen{\textbf{11.5}}}}$ & 58.2$_{\mathclap{\hspace{1.2em}\small\darkred{\textbf{-1.3}}}}$ \\
     & COIECD & 59.9$_{\mathclap{\hspace{0.9em}\small\darkgreen{\textbf{8.0}}}}$ & 76.0$_{\mathclap{\hspace{0.9em}\small\darkgreen{\textbf{4.4}}}}$ &
     77.8$_{\mathclap{\hspace{1.2em}\small\darkred{\textbf{-3.6}}}}$ &48.9$_{\mathclap{\hspace{1.2em}\small\darkgreen{\textbf{12.4}}}}$ & 48.4$_{\mathclap{\hspace{1.2em}\small\darkgreen{\textbf{14.6}}}}$ &75.9$_{\mathclap{\hspace{0.9em}\small\darkgreen{\textbf{9.9}}}}$  &64.5$_{\mathclap{\hspace{0.9em}\small\darkgreen{\textbf{5.6}}}}$  \\
     & ADACAD & 65.4$_{\mathclap{\hspace{1.2em}\small\darkgreen{\textbf{13.5}}}}$ & 74.6$_{\mathclap{\hspace{0.9em}\small\darkgreen{\textbf{3.0}}}}$ &
     82.3$_{\mathclap{\hspace{0.9em}\small\darkgreen{\textbf{0.9}}}}$ &46.1$_{\mathclap{\hspace{0.9em}\small\darkgreen{\textbf{9.6}}}}$ & 44.3$_{\mathclap{\hspace{1.2em}\small\darkgreen{\textbf{10.5}}}}$ & 72.0$_{\mathclap{\hspace{0.9em}\small\darkgreen{\textbf{6.0}}}}$ &64.1$_{\mathclap{\hspace{0.9em}\small\darkgreen{\textbf{5.2}}}}$  \\
    \cmidrule(){2-9}
     & \cellcolor[HTML]{FDF5F1}\textbf{\ourapproach(Ours)} & \cellcolor[HTML]{FDF5F1} \textbf{68.4}$_{\mathclap{\hspace{1.2em}\small\darkgreen{\textbf{16.5}}}}$ & \cellcolor[HTML]{FDF5F1}\textbf{83.2}$_{\mathclap{\hspace{1.2em}\small\darkgreen{\textbf{11.6}}}}$& \cellcolor[HTML]{FDF5F1}\textbf{83.9}$_{\mathclap{\hspace{0.9em}\small\darkgreen{\textbf{2.5}}}}$ 
     & \cellcolor[HTML]{FDF5F1}\textbf{54.2}$_{\mathclap{\hspace{1.2em}\small\darkgreen{\textbf{17.7}}}}$ & \cellcolor[HTML]{FDF5F1}\textbf{57.4}$_{\mathclap{\hspace{1.2em}\small\darkgreen{\textbf{23.6}}}}$& \cellcolor[HTML]{FDF5F1}\textbf{81.1}$_{\mathclap{\hspace{1.2em}\small\darkgreen{\textbf{15.1}}}}$ & \cellcolor[HTML]{FDF5F1}\textbf{71.4}$_{\mathclap{\hspace{1.2em}\small\darkgreen{\textbf{14.5}}}}$ \\
    \midrule
    \midrule
    \multirow{5}{*}{Llama2-13B} 
     & \cellcolor[gray]{0.945}Greedy & \cellcolor[gray]{0.945}64.3 & \cellcolor[gray]{0.945}73.9 & \cellcolor[gray]{0.945}\textbf{86.1} & \cellcolor[gray]{0.945}36.4 & \cellcolor[gray]{0.945}53.0 & \cellcolor[gray]{0.945}73.7 &  \cellcolor[gray]{0.945}64.6\\
    \cmidrule(){2-9}
     & CAD & 51.3$_{\mathclap{\hspace{1.5em}\small\darkred{\textbf{-13.0}}}}$ & 72.7$_{\mathclap{\hspace{1.2em}\small\darkred{\textbf{-1.2}}}}$ & 48.9$_{\mathclap{\hspace{1.5em}\small\darkred{\textbf{-37.2}}}}$ &  53.9$_{\mathclap{\hspace{1.2em}\small\darkgreen{\textbf{17.5}}}}$ & 62.3$_{\mathclap{\hspace{0.9em}\small\darkgreen{\textbf{9.3}}}}$ & 80.1$_{\mathclap{\hspace{0.9em}\small\darkgreen{\textbf{6.4}}}}$ & 61.5$_{\mathclap{\hspace{1.2em}\small\darkred{\textbf{-3.1}}}}$ \\
     & COIECD & 68.6$_{\mathclap{\hspace{0.9em}\small\darkgreen{\textbf{4.3}}}}$ & 78.9$_{\mathclap{\hspace{0.9em}\small\darkgreen{\textbf{5.0}}}}$ &85.8$_{\mathclap{\hspace{1.2em}\small\darkred{\textbf{-0.3}}}}$ & 54.0$_{\mathclap{\hspace{1.2em}\small\darkgreen{\textbf{17.6}}}}$ & 56.8$_{\mathclap{\hspace{0.9em}\small\darkgreen{\textbf{3.8}}}}$ & 76.8$_{\mathclap{\hspace{0.9em}\small\darkgreen{\textbf{3.1}}}}$ &70.2$_{\mathclap{\hspace{0.9em}\small\darkgreen{\textbf{5.6}}}}$    \\
     & ADACAD & 68.2$_{\mathclap{\hspace{0.9em}\small\darkgreen{\textbf{3.9}}}}$ & 76.5$_{\mathclap{\hspace{0.9em}\small\darkgreen{\textbf{2.6}}}}$ & 85.7$_{\mathclap{\hspace{1.2em}\small\darkred{\textbf{-0.4}}}}$ & 62.6$_{\mathclap{\hspace{1.2em}\small\darkgreen{\textbf{26.2}}}}$ & 62.5$_{\mathclap{\hspace{0.9em}\small\darkgreen{\textbf{9.5}}}}$ & 77.4$_{\mathclap{\hspace{0.9em}\small\darkgreen{\textbf{3.7}}}}$ & 72.2$_{\mathclap{\hspace{0.9em}\small\darkgreen{\textbf{7.6}}}}$  \\
    \cmidrule(){2-9}
     & \cellcolor[HTML]{FDF5F1}\textbf{\ourapproach(Ours)} & \cellcolor[HTML]{FDF5F1}\textbf{71.4}$_{\mathclap{\hspace{0.9em}\small\darkgreen{\textbf{7.1}}}}$ & \cellcolor[HTML]{FDF5F1}\textbf{79.5}$_{\mathclap{\hspace{1.2em}\small\darkgreen{\textbf{5.6}}}}$& \cellcolor[HTML]{FDF5F1}\textbf{86.1}$_{\mathclap{\hspace{0.9em}\small\darkgreen{\textbf{0.0}}}}$ & \cellcolor[HTML]{FDF5F1}\textbf{65.5}$_{\mathclap{\hspace{1.2em}\small\darkgreen{\textbf{29.1}}}}$ &\cellcolor[HTML]{FDF5F1}\textbf{65.2}$_{\mathclap{\hspace{1.2em}\small\darkgreen{\textbf{12.2}}}}$  & \cellcolor[HTML]{FDF5F1}\textbf{86.0}$_{\mathclap{\hspace{1.2em}\small\darkgreen{\textbf{12.3}}}}$ &  \cellcolor[HTML]{FDF5F1}\textbf{75.6}$_{\mathclap{\hspace{1.2em}\small\darkgreen{\textbf{11.0}}}}$\\
    \midrule
    \midrule
    \multirow{5}{*}{Llama3-8B} 
     & \cellcolor[gray]{0.945}Greedy & \cellcolor[gray]{0.945}67.4 & \cellcolor[gray]{0.945}87.2&\cellcolor[gray]{0.945}\textbf{88.6}& \cellcolor[gray]{0.945}47.2 & \cellcolor[gray]{0.945}48.1& \cellcolor[gray]{0.945}67.9 & \cellcolor[gray]{0.945}67.7 \\
    \cmidrule(){2-9}
     & CAD & 51.0$_{\mathclap{\hspace{1.5em}\small\darkred{\textbf{-16.4}}}}$ & 72.6$_{\mathclap{\hspace{1.5em}\small\darkred{\textbf{-14.6}}}}$ & 55.8$_{\mathclap{\hspace{1.5em}\small\darkred{\textbf{-32.8}}}}$& 56.0$_{\mathclap{\hspace{0.9em}\small\darkgreen{\textbf{8.8}}}}$ & 57.6$_{\mathclap{\hspace{0.9em}\small\darkgreen{\textbf{9.5}}}}$ & 68.4$_{\mathclap{\hspace{1.2em}\small\darkred{\textbf{-4.7}}}}$ & 
     60.2$_{\mathclap{\hspace{1.2em}\small\darkred{\textbf{-7.5}}}}$\\
     & COIECD & 68.4$_{\mathclap{\hspace{0.9em}\small\darkgreen{\textbf{1.0}}}}$ & 86.9$_{\mathclap{\hspace{1.2em}\small\darkred{\textbf{-0.3}}}}$ & 87.5$_{\mathclap{\hspace{1.2em}\small\darkred{\textbf{-1.1}}}}$ & 52.1$_{\mathclap{\hspace{0.9em}\small\darkgreen{\textbf{4.9}}}}$ & 52.1$_{\mathclap{\hspace{0.9em}\small\darkgreen{\textbf{4.0}}}}$ & 70.3$_{\mathclap{\hspace{0.9em}\small\darkgreen{\textbf{2.4}}}}$ & 69.9$_{\mathclap{\hspace{0.9em}\small\darkgreen{\textbf{2.2}}}}$ \\
     & ADACAD & 65.2$_{\mathclap{\hspace{1.2em}\small\darkred{\textbf{-2.2}}}}$ & 87.1$_{\mathclap{\hspace{1.2em}\small\darkred{\textbf{-0.1}}}}$& 86.3$_{\mathclap{\hspace{1.2em}\small\darkred{\textbf{-2.3}}}}$ & 58.3$_{\mathclap{\hspace{1.2em}\small\darkgreen{\textbf{11.1}}}}$ & 59.0$_{\mathclap{\hspace{1.2em}\small\darkgreen{\textbf{10.9}}}}$ & 77.5$_{\mathclap{\hspace{0.9em}\small\darkgreen{\textbf{9.6}}}}$ & 71.0$_{\mathclap{\hspace{0.9em}\small\darkgreen{\textbf{3.3}}}}$ \\
    \cmidrule(){2-9}
     & \cellcolor[HTML]{FDF5F1}\textbf{\ourapproach(Ours)} & \cellcolor[HTML]{FDF5F1}\textbf{73.4}$_{\mathclap{\hspace{0.9em}\small\darkgreen{\textbf{6.0}}}}$ &  \cellcolor[HTML]{FDF5F1}\textbf{88.9}$_{\mathclap{\hspace{0.9em}\small\darkgreen{\textbf{1.7}}}}$& \cellcolor[HTML]{FDF5F1}88.0$_{\mathclap{\hspace{1.2em}\small\darkred{\textbf{-0.6}}}}$ & \cellcolor[HTML]{FDF5F1}\textbf{65.3}$_{\mathclap{\hspace{1.2em}\small\darkgreen{\textbf{18.1}}}}$ & \cellcolor[HTML]{FDF5F1}\textbf{67.3}$_{\mathclap{\hspace{1.2em}\small\darkgreen{\textbf{19.2}}}}$  & \cellcolor[HTML]{FDF5F1}\textbf{79.5}$_{\mathclap{\hspace{1.2em}\small\darkgreen{\textbf{11.6}}}}$ &\cellcolor[HTML]{FDF5F1}\textbf{77.1}$_{\mathclap{\hspace{0.9em}\small\darkgreen{\textbf{9.4}}}}$ \\
    \midrule
    \midrule
    \multirow{5}{*}{Mistral-7B} 
     & \cellcolor[gray]{0.945}Greedy & \cellcolor[gray]{0.945}58.8 & \cellcolor[gray]{0.945}67.5& \cellcolor[gray]{0.945}75.3 & \cellcolor[gray]{0.945}45.6 & \cellcolor[gray]{0.945}41.6& \cellcolor[gray]{0.945}66.2 &\cellcolor[gray]{0.945}59.2  \\
    \cmidrule(){2-9}
     & CAD & 52.6$_{\mathclap{\hspace{1.2em}\small\darkred{\textbf{-6.2}}}}$ & 25.2$_{\mathclap{\hspace{1.5em}\small\darkred{\textbf{-42.3}}}}$ & 23.3$_{\mathclap{\hspace{1.5em}\small\darkred{\textbf{-52.0}}}}$& 48.1$_{\mathclap{\hspace{0.9em}\small\darkgreen{\textbf{2.5}}}}$ & 45.8$_{\mathclap{\hspace{0.9em}\small\darkgreen{\textbf{4.2}}}}$ & 36.3$_{\mathclap{\hspace{1.5em}\small\darkred{\textbf{-29.9}}}}$ & 38.6$_{\mathclap{\hspace{1.5em}\small\darkred{\textbf{-20.6}}}}$ \\
     & COIECD & 59.1$_{\mathclap{\hspace{0.9em}\small\darkgreen{\textbf{0.3}}}}$ & 48.1$_{\mathclap{\hspace{1.5em}\small\darkred{\textbf{-19.4}}}}$ & 58.6$_{\mathclap{\hspace{1.5em}\small\darkred{\textbf{-16.7}}}}$& 56.0$_{\mathclap{\hspace{1.2em}\small\darkgreen{\textbf{10.4}}}}$ & 54.0$_{\mathclap{\hspace{1.2em}\small\darkgreen{\textbf{12.4}}}}$ & 69.1$_{\mathclap{\hspace{0.9em}\small\darkgreen{\textbf{2.9}}}}$ & 
     57.5$_{\mathclap{\hspace{1.2em}\small\darkred{\textbf{-1.7}}}}$\\
     & ADACAD & 58.1$_{\mathclap{\hspace{1.2em}\small\darkred{\textbf{-0.7}}}}$ & \textbf{75.2}$_{\mathclap{\hspace{0.9em}\small\darkgreen{\textbf{7.7}}}}$ &74.7$_{\mathclap{\hspace{1.2em}\small\darkred{\textbf{-0.6}}}}$  & 51.7$_{\mathclap{\hspace{0.9em}\small\darkgreen{\textbf{6.1}}}}$ & 49.8$_{\mathclap{\hspace{0.9em}\small\darkgreen{\textbf{8.2}}}}$ & 67.4$_{\mathclap{\hspace{0.9em}\small\darkgreen{\textbf{1.2}}}}$ & 62.8$_{\mathclap{\hspace{1.2em}\small\darkgreen{\textbf{3.6}}}}$\\
    \cmidrule(){2-9}
     & \cellcolor[HTML]{FDF5F1}\textbf{\ourapproach(Ours)} & \cellcolor[HTML]{FDF5F1}\textbf{60.9}$_{\mathclap{\hspace{0.9em}\small\darkgreen{\textbf{2.1}}}}$ & \cellcolor[HTML]{FDF5F1}69.1$_{\mathclap{\hspace{0.9em}\small\darkgreen{\textbf{1.6}}}}$  & \cellcolor[HTML]{FDF5F1}\textbf{77.6}$_{\mathclap{\hspace{0.9em}\small\darkgreen{\textbf{2.3}}}}$& \cellcolor[HTML]{FDF5F1}\textbf{58.3}$_{\mathclap{\hspace{1.2em}\small\darkgreen{\textbf{12.7}}}}$ &\cellcolor[HTML]{FDF5F1}\textbf{57.9}$_{\mathclap{\hspace{1.2em}\small\darkgreen{\textbf{16.3}}}}$  & \cellcolor[HTML]{FDF5F1}\textbf{72.6}$_{\mathclap{\hspace{0.9em}\small\darkgreen{\textbf{6.4}}}}$ &\cellcolor[HTML]{FDF5F1}\textbf{66.1}$_{\mathclap{\hspace{0.9em}\small\darkgreen{\textbf{6.9}}}}$  \\
    \bottomrule
    \end{tabular}
}

\label{tab:main_qa}
\end{table*}

\subsection{Experimental Setup}
\subsubsection{Datasets}
We conducted extensive evaluations of \ourapproach using both standard question-answering datasets: Natural Questions (NQ)~\cite{kwiatkowski-etal-2019-natural}, TriviaQA~\cite{joshi2017triviaqa} and SQuAD~\cite{rajpurkar2016squad}, as well as knowledge conflict question-answering datasets: Counterfacts~\cite{longpre2022entitybasedknowledgeconflictsquestion} and NQ-Swap~\cite{longpre2022entitybasedknowledgeconflictsquestion}.
Additionally, as described in Section~\ref{subs:ConlictQA} we have developed a new benchmark, ConflictKG, employing a generative approach to simulate real-world knowledge conflicts.

\textbf{Natural Questions (NQ)} is a large-scale question answering corpus comprising real user queries from Google search, paired with answers from Wikipedia. It includes 307,373 training examples, 7,830 development examples, and 7,842 test examples, annotated for long and short answers. The dataset is designed to evaluate QA systems, with robust metrics and high human performance baselines. A subset of 3,231 validation instances with short answers is often used for benchmarking. 

\textbf{NQ-Swap} is a dataset designed to evaluate models' ability to handle knowledge conflicts. It creates synthetic conflicts by swapping named entities in the context, challenging models to prioritize contextual over parametric knowledge. Derived from the NQ dataset, it consists of 4K instances, aiding in assessing and mitigating over-reliance on memorized information.

\textbf{SQuAD} is a reading comprehension dataset with over 100,000 questions created by crowdworkers on Wikipedia articles. Each question's answer is a text segment from the corresponding passage. The dataset requires various reasoning skills, analyzed using dependency and constituency trees.

\textbf{TriviaQA} is a challenging reading comprehension dataset containing over 650K question-answer-evidence triples. TriviaQA includes 95K question-answer pairs authored by trivia enthusiasts and independently gathered evidence documents, six per question on average, that provide high quality distant supervision for answering the questions. 


\textbf{ConflictKG} is a knowledge conflict benchmark that simulates real-world scenarios, containing 4,466 conflict and non-conflict instances along with their knowledge sources. The construction process can be referred to in Section~\ref{subs:ConlictQA}.

\subsubsection{LLM \& Baselines}
We conducted experiments on four open-source LLMs, including Llama2-7b~\cite{touvron2023llama2openfoundation}, Llama2-13b~\cite{touvron2023llama2openfoundation}, Llama3-8b~\cite{grattafiori2024llama3herdmodels}, and Mistral-7b~\cite{jiang2023mistral7b}.
And we consider four decoding strategies as baselines, including: Context-aware decoding (CAD)~\cite{shi2023trustingevidencehallucinatecontextaware}, COIECD~\cite{yuan2024discerningresolvingknowledgeconflicts} and ADACAD~\cite{wang2024adacadadaptivelydecodingbalance}.
Greedy decoding is the standard decoding strategy. CAD mitigates knowledge conflicts by comparing the output distributions with and without context, thereby controlling the decoding process. COIECD and ADACAD are two dynamic decoding strategies. COIECD adjusts the decoding process by imposing constraints based on the entropy of contextual information, while ADACAD regulates the process by calculating the Jensen-Shannon Divergence between the output distributions with and without context.

\subsubsection{Implementation Details}
To ensure a fair comparison, we standardized the sampling hyperparameters across \ourapproach and all baseline methods, using the simplest zero-shot prompt template.
For CAD, we set $\alpha = 1$; for COIECD, we set $\lambda$ and $\alpha$ to 0.25 and 1, respectively; for \ourapproach, we set both $\lambda$ and $\alpha$ to 1.

To ensure fairness in the decoding process, we use a consistent zero-shot template for all baselines: 
``\{context\}\textbackslash n Using only the references listed above, answer the following question: \textbackslash n Question: \{question\}\textbackslash n Answer''.
During inference, the question $q$ and context $c$ will be inserted into the corresponding places in the template. Additionally, we set the maximum generation length to 32 for all methods to avoid the impact of varying decoding depths on the evaluation results.

\subsubsection{Evaluation Metrics}
Traditional exact match (EM) methods are increasingly insufficient for complex context-based question answering tasks. Given the growing adoption of large language models (LLMs) for answer quality evaluation, we developed an automated evaluation framework, designed a prompt-based evaluation template, and employed GPT-4o\footnote{GPT-4o is from \url{https://openai.com/}} to evaluate the answers.

We adopt a generative approach to evaluate open-domain QA tasks, aiming to overcome the limitations of traditional evaluation methods. The traditional EM evaluation primarily relies on exact matching between the generated answer and the reference answer. However, in many real-world scenarios, especially with open-ended questions, EM matching no longer fully reflects the model's true performance. Even if the generated answer is not an exact match with the reference answer, it can still be considered correct as long as the meaning and logic align.
By using generative models like GPT for evaluation, we can provide a more flexible and human-like assessment of the answers, offering a more accurate measure of the model's actual capabilities. The specific prompt template is shown in Table~\ref{tab:prompteval}.

\subsection{Main Results}

\paragraph{Results on Knowledge Conflict QA benchmark}
We conducted experiments on Counterfacts and NQ-Swap, simulating high-conflict contextual environments, to evaluate the handling capacity of \ourapproach for high-conflict information.
The results, shown in Table~\ref{tab:main_qa}, demonstrate that \ourapproach consistently outperforms greedy decoding, CAD, COIECD, and ADACAD.
For example, on NQ-Swap, \ourapproach outperforms the baseline greedy decoding by 17.7\%, 29.1\%, 12.7\%, and 18.1\% on Llama2-7b, Llama2-13b, Llama3-8b, and Mistral-7b, respectively. It also exceeds the baseline CAD by 6.7\%, 11.6\%, 10.2\%, and 9.3\%, respectively.
In scenarios with frequent knowledge conflicts, \ourapproach achieves significant improvements. This highlights the effectiveness of our dynamic decoding strategy, which adjusts intervention strength based on the level of conflict, thereby enhancing the model’s ability to manage complex conflict scenarios.

\paragraph{Results on General QA benchmark}
We conducted experiments on NQ, SQuAD and TriviaQA, simulating low-conflict contexts, to evaluate the handling capacity of \ourapproach for low-conflict information. The results, presented in Table~\ref{tab:main_qa}, show that \ourapproach consistently outperforms all baselines.
For example, on the NQ dataset, \ourapproach surpasses the baseline greedy decoding by 16.5\%, 7.1\%, 2.1\%, and 6.0\% on Llama2-7b, Llama2-13b, Llama3-8b, and Mistral-7b, respectively. It also outperforms the baseline CAD by 18.1\%, 20.1\%, 8.3\%, and 22.4\%, respectively.
It is noteworthy that CAD underperforms compared to the baseline greedy decoding in low-conflict scenarios, with an average accuracy drop of 8.1\% across all models. In contrast, \ourapproach consistently surpasses all baselines in every low-conflict scenario.
This clearly highlights the superiority of our approach, which employs a conflict prediction mechanism to route conflicting and non-conflicting information along distinct decoding paths.
Meanwhile, \ourapproach surpasses the two dynamic contrastive decoding methods, ADACAD and COIECD, underscoring the rationale and effectiveness of dynamically adjusting the decoding process based on contextual fidelity.

\begin{table*}[t]
\small
\renewcommand{\arraystretch}{1.30}
\centering
\belowrulesep=0pt
\aboverulesep=0pt
\caption{We conduct experiments on ConflictKG to analyze the performance of five decoding methods in both conflict and non-conflict scenarios.The black values denote each model's QA accuracy on conflict and non-conflict data across different methods. The subscripts represent the performance difference between the current method and the baseline, where red indicates performance degradation and green indicates performance improvement.}
\begin{tabular}{c|cc|cc|cc|cc}
\toprule
\rowcolor[gray]{0.9}
 & \multicolumn{2}{c|}{\textbf{Llama2-7b}}  & \multicolumn{2}{c|}{\textbf{Llama-13b}} & \multicolumn{2}{c|}{\textbf{Llama3-8b}} & \multicolumn{2}{c}{\textbf{Mistral-7b}} \\ 
 \rowcolor[gray]{0.9}
\multirow{-2}{*}{\textbf{Decoding}} & \textbf{conflict} & \textbf{non conflict} & \textbf{conflict} & \textbf{non conflict} & \textbf{conflict} & \textbf{non conflict} & \textbf{conflict} & \textbf{non conflict} \\ \midrule
\rowcolor[gray]{0.945}
Greedy & 53.8 & 78.1 & 65.7 & 81.5 & 58.3 & 77.5 & 56.4 & 75.9 \\
CAD & 72.9$_{\mathclap{\hspace{1.2em}\small\darkgreen{\textbf{19.1}}}}$ & 82.0$_{\mathclap{\hspace{0.9em}\small\darkgreen{\textbf{3.9}}}}$ & 77.9$_{\mathclap{\hspace{1.2em}\small\darkgreen{\textbf{12.2}}}}$ & 83.1$_{\mathclap{\hspace{0.9em}\small\darkgreen{\textbf{1.6}}}}$ & 66.1$_{\mathclap{\hspace{0.9em}\small\darkgreen{\textbf{7.8}}}}$ & 70.7$_{\mathclap{\hspace{1.2em}\small\darkred{\textbf{-6.8}}}}$ & 49.7$_{\mathclap{\hspace{1.2em}\small\darkred{\textbf{-6.7}}}}$ & 22.9$_{\mathclap{\hspace{1.5em}\small\darkred{\textbf{-53.0}}}}$ \\
COIECD & 68.8$_{\mathclap{\hspace{1.2em}\small\darkgreen{\textbf{15.0}}}}$ & 82.9$_{\mathclap{\hspace{0.9em}\small\darkgreen{\textbf{4.8}}}}$ & 67.7$_{\mathclap{\hspace{0.9em}\small\darkgreen{\textbf{2.0}}}}$ & 85.8$_{\mathclap{\hspace{0.9em}\small\darkgreen{\textbf{4.3}}}}$ & 67.2$_{\mathclap{\hspace{0.9em}\small\darkgreen{\textbf{8.9}}}}$ & 77.6$_{\mathclap{\hspace{0.9em}\small\darkgreen{\textbf{0.1}}}}$ & 67.9$_{\mathclap{\hspace{1.2em}\small\darkgreen{\textbf{11.5}}}}$ & 70.3$_{\mathclap{\hspace{1.2em}\small\darkred{\textbf{-5.6}}}}$ \\
ADACAD & 61.4$_{\mathclap{\hspace{0.9em}\small\darkgreen{\textbf{7.6}}}}$ & 82.7$_{\mathclap{\hspace{0.9em}\small\darkgreen{\textbf{9.2}}}}$ & 70.7$_{\mathclap{\hspace{0.9em}\small\darkgreen{\textbf{5.0}}}}$ & 84.1$_{\mathclap{\hspace{0.9em}\small\darkgreen{\textbf{2.6}}}}$ & 66.2$_{\mathclap{\hspace{0.9em}\small\darkgreen{\textbf{7.9}}}}$ & 74.4$_{\mathclap{\hspace{1.2em}\small\darkred{\textbf{-3.1}}}}$ & 59.4$_{\mathclap{\hspace{0.9em}\small\darkgreen{\textbf{3.0}}}}$ & 75.4$_{\mathclap{\hspace{1.2em}\small\darkred{\textbf{-0.5}}}}$ \\
\rowcolor[HTML]{FDF5F1}
\textbf{\ourapproach(Ours)} & \textbf{74.8}$_{\mathclap{\hspace{1.2em}\small\darkgreen{\textbf{21.0}}}}$ & \textbf{87.3}$_{\mathclap{\hspace{0.9em}\small\darkgreen{\textbf{9.2}}}}$ & \textbf{82.5}$_{\mathclap{\hspace{1.2em}\small\darkgreen{\textbf{16.8}}}}$ & \textbf{89.5}$_{\mathclap{\hspace{0.9em}\small\darkgreen{\textbf{8.0}}}}$ & \textbf{73.8}$_{\mathclap{\hspace{1.2em}\small\darkgreen{\textbf{15.5}}}}$ & \textbf{85.0}$_{\mathclap{\hspace{0.9em}\small\darkgreen{\textbf{7.5}}}}$ & \textbf{68.0}$_{\mathclap{\hspace{1.2em}\small\darkgreen{\textbf{11.6}}}}$ & \textbf{77.1}$_{\mathclap{\hspace{0.9em}\small\darkgreen{\textbf{1.2}}}}$ \\ \bottomrule
\end{tabular}

\label{tab:conf_QA}
 \vspace{-6mm}
\end{table*}

\paragraph{Results on ConflictKG}
We conducted experiments on ConflictKG, simulating complex scenarios involving context-memory conflicts, which commonly arise in situations where knowledge frequently updates and model updates lag behind. The experimental results, presented in Table~\ref{tab:main_qa}, show that \ourapproach outperforms all baselines.
The average results of \ourapproach across the four LLMs exceed baseline greedy decoding, CAD, COIECD, and ADACAD by 11.3\%, 14.2\%, 6.8\%, and 6.2\%, respectively.
This indicates that \ourapproach can emulate how the brain selectively focuses and dynamically adjusts its decision-making process when confronted with conflicting information, flexibly determining when correction is necessary, rather than indiscriminately processing all data. As a result, it maintains high robustness even in complex conflict scenarios.
Notably, \ourapproach intelligently selects decoding paths based on the level of conflict. Unlike other dynamic decoding methods, it minimizes unnecessary computational overhead, thereby improving reasoning efficiency and accuracy, while showcasing its distinctive advantages in complex scenarios.

\begin{figure}
	\centering
	\includegraphics[width=1.0\linewidth]{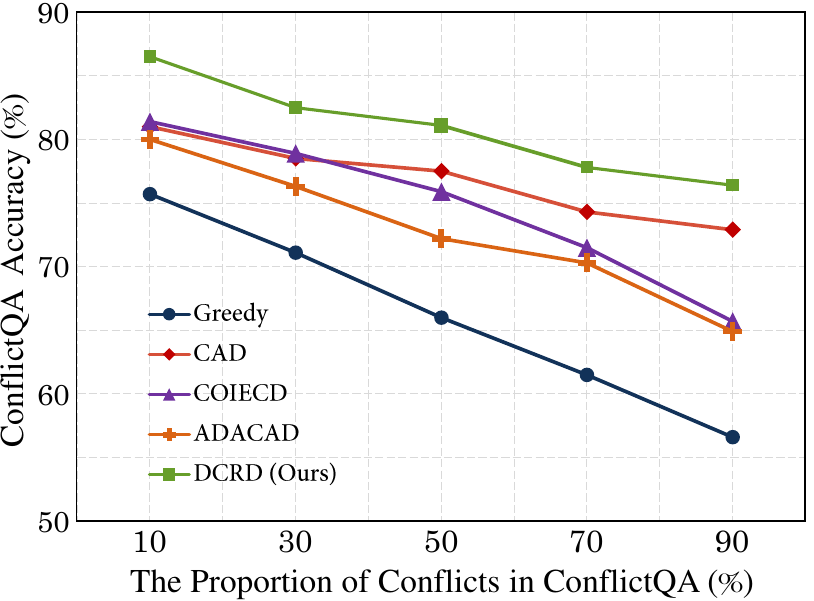}
	\caption{We experiment on ConflictKG by varying the conflict data proportion. \ourapproach consistently outperforms other baselines, with smaller fluctuations in performance as the conflict ratio changes.}
	\label{fig:zhexian1}
        \vspace{-4mm}
\end{figure}

\subsection{Ablation Study}

\subsubsection{Role of Cognitive Reconciliation Decoding}
To better isolate the contribution of the Cognitive Reconciliation Decoding (CRD) module, we conducted an ablation study in which the dynamic routing mechanism was removed. In this setting, the CRD strategy is applied to all inputs uniformly, regardless of whether a knowledge conflict is present. This configuration allows us to directly assess the intrinsic impact of the CRD mechanism on generation quality.

By comparing this variant against a range of existing contrastive decoding approaches (e.g., CAD, COIECD, and ADACAD), we are able to objectively evaluate the standalone effectiveness of CRD, as shown in Table~\ref{tab:crd_abla}.

The results demonstrate that even in the absence of the routing module, our cognitive reconciliation decoding strategy achieves leading performance on both conflict and non-conflict samples. This finding suggests that the observed performance improvements are not solely attributable to dynamic routing. Instead, it highlights that the decoding strategy itself contributes significantly to improving controllability in language generation.

\subsubsection{Role of Conflict Predictor}
As shown in Table~\ref{tab:layer}, we conducted a comparative experiment on ConflictKG using Llama2-7b with different conflict predictor settings. ``Random'' refers to randomly generated classification results.``Result'' refers to the accuracy achieved by applying the corresponding classifier settings on ConflictKG. The red subscript represents the difference from our method.
First, with randomly generated classification results (accuracy of 50\%), the performance of the QA task dropped significantly by 9.6\%. For the classifier based on hidden states, we conducted tests at the 16th and 32nd layers. The results revealed that the classifier's accuracy was 8.2\% and 11.5\% lower than our method, resulting in performance drops of 3.3\% and 4\%, respectively. These findings suggest that the conflict predictor plays a crucial role in maintaining the overall performance of the system.

\begin{table}[t]
    \centering
    \small
     \caption{
    In this ablation experiment, we removed the dynamic routing (DR) mechanism and applied the CRD strategy to all inputs, regardless of whether a conflict exists.
    }
    \belowrulesep=0pt
    \aboverulesep=0pt
    \setlength\tabcolsep{10pt}
    \renewcommand\arraystretch{1.3}
    \resizebox{1\columnwidth}{!}{%
    \setlength{\tabcolsep}{8pt} 
    
        \begin{tabular}{c|cc|c}
        \toprule
        \rowcolor[gray]{0.9}
        \textbf{Method}
        & \textbf{Conflict} & \textbf{Non-Conflict} & \textbf{Total} \\ 

        \hline
        \rowcolor[gray]{0.945}
        Greedy & 53.8 & 78.1 & 66.0  \\
        \hline
        CAD & 72.9 & 82.0 & 77.5  \\
        \hline
        COIECD & 68.8 & 82.9 & 75.9  \\
        \hline
        ADACAD & 61.4 & 82.7 & 72.0  \\
        \hline
        \rowcolor[HTML]{FDF5F1}
        \textbf{\ourapproach w/o DR} & \textbf{73.9} &\textbf{83.2} & \textbf{78.6}\\
        
        \bottomrule[0.4pt]
        \end{tabular}%
    }

    \label{tab:crd_abla}
    \vspace{-4mm}
\end{table}

\begin{table}[t]
    \centering
    \Large
     \caption{
    Performance of Llama2-7b when trained across different datasets.Original Acc denotes the accuracy when the test set is consistent with the training set, while Final Acc. represents the accuracy when a different test set is used.
    }
    \belowrulesep=0pt
    \aboverulesep=0pt
    \setlength\tabcolsep{10pt}
    \renewcommand\arraystretch{1.3}
    \resizebox{1\columnwidth}{!}{%
    \setlength{\tabcolsep}{8pt} 
    
        \begin{tabular}{cccc}
        \toprule
        \rowcolor[gray]{0.9}
        \textbf{Training Data} & \textbf{Test Data} & \textbf{Original Acc.} & \textbf{Final Acc.} \\ 
        \hline
        \multicolumn{4}{c}{\textit{NQ \& NQ-swap }} \\
        \hline
        NQ-swap & NQ & 68.4 & 67.1$_{\mathclap{\hspace{1.2em}\small\darkred{\textbf{-1.3}}}}$  \\
        \hline
        NQ & NQ-swap & 54.2 & 51.1$_{\mathclap{\hspace{1.2em}\small\darkred{\textbf{-3.1}}}}$  \\
        \hline
        NQ-swap + NQ & NQ & 68.4 & 68.7$_{\mathclap{\hspace{1.2em}\small\darkgreen{\textbf{+0.3}}}}$  \\
        \hline
        NQ-swap + NQ & NQ-swap& 54.2 & 53.9$_{\mathclap{\hspace{1.2em}\small\darkred{\textbf{-0.3}}}}$   \\
        \hline
        \multicolumn{4}{c}{\textit{ NQ-swap \& ConflictKG }} \\
        \hline
        NQ-swap & ConflictKG & 81.1 & 76.9$_{\mathclap{\hspace{1.2em}\small\darkred{\textbf{-4.2}}}}$  \\
        \hline
        ConflictKG & NQ-swap & 54.2 & 49.9$_{\mathclap{\hspace{1.2em}\small\darkred{\textbf{-4.3}}}}$  \\
        \hline
        NQ-swap + ConflictKG & ConflictKG & 81.1 & 78.1$_{\mathclap{\hspace{1.2em}\small\darkred{\textbf{-3.0}}}}$  \\
        \hline
        NQ-swap + ConflictKG & NQ-swap& 54.2 & 53.6$_{\mathclap{\hspace{1.2em}\small\darkred{\textbf{-0.6}}}}$   \\

        \bottomrule[0.4pt]
        \end{tabular}%
    }

    \label{tab:data_trans}
    \vspace{-4mm}
\end{table}

\subsection{Analysis}

\begin{table}[t]
    \centering
    \footnotesize  

     \caption{
        The Impact of Different Classifier Settings on Llama2-7b.Results denote the final question-answering accuracy obtained by using different conflict predictors.
    }
    \belowrulesep=0pt
    \aboverulesep=0pt
    \setlength\tabcolsep{8pt}
    \renewcommand\arraystretch{1.0}
    
    \resizebox{0.9\columnwidth}{!}{%
    \setlength{\tabcolsep}{8pt} 
    
        \begin{tabular}{ccc}
        \toprule
        \rowcolor[gray]{0.9}
        \multicolumn{1}{c}{\multirow{1}{*}{\textbf{Layer}}} 
        & \textbf{Conflict Predictor} & \textbf{Results} \\ 
        \hline
        \multicolumn{3}{c}{\textit{Random}} \\
        \hline

        - & 50 & 71.5$_{\mathclap{\hspace{1.2em}\small\darkred{\textbf{-9.6}}}}$  \\
 
        \hline
        \multicolumn{3}{c}{\textit{Hidden state}} \\
        \hline
           16th Layer & 76.5 & 77.8$_{\mathclap{\hspace{1.2em}\small\darkred{\textbf{-4.6}}}}$ \\
           32nd Layer & 73.2 & 77.1$_{\mathclap{\hspace{1.2em}\small\darkred{\textbf{-4.0}}}}$ \\
       
        \hline
        \multicolumn{3}{c}{\textit{Attention maps (Ours)}} \\
        \hline
        \rowcolor[HTML]{FDF5F1}
        32 Layers & 84.7 & 81.1 \\
       
        \hline
        
        \bottomrule[0.4pt]
        \end{tabular}%
    }

    \label{tab:layer}

\end{table}

\begin{table}[t]
    \centering
    \footnotesize 
    
    \caption{
        Comparison of Inference Time and Average Performance.
    }
    
    \belowrulesep=0pt
    \aboverulesep=0pt
    \setlength\tabcolsep{8pt}
    \renewcommand\arraystretch{1.1} 
    
    \resizebox{\columnwidth}{!}{%
    \setlength{\tabcolsep}{12pt} 
    
        \begin{tabular}{ccc}
        \toprule
        \textbf{Method} & \textbf{Inference time (s)} & \textbf{Avg. Performance} \\ 
        \hline
        
        CAD & \textbf{0.50} & 58.2 \\
        ADACAD & 0.52 & 64.1 \\
        COIECD & 2.4 & 64.5 \\
        
        \hline
        \textbf{DCRD (ours)} & 0.96 & \textbf{71.4} \\
        
        \bottomrule[0.4pt]
        \end{tabular}%
    }

    \label{tab:time_performance_comparison}
\end{table}

\begin{table*}[!t]
\small
\centering
\caption{A case study using Llama2-7b on ConflictKG. \textcolor{darkgreen}{Green} text indicates the correct answer, \textcolor{darkgolden}{yellow} text indicates a partially correct answer, and \textcolor{darkred}{red} text indicates an incorrect answer.
}
\begin{tabular}{l}
\toprule
\addlinespace[4pt]
\parbox[c]{15.6cm}{
\textbf{Context:}  The Super Bowl 50 Halftime Show took place on February 7, 2016, at Levi's Stadium in Santa Clara, California as part of Super Bowl 50 . It was headlined by the British rock group \textcolor{darkgreen}{Coldplay with special guest performers Beyoncé and Bruno Mars}  , who previously had headlined the Super Bowl XLVII and Super Bowl XLVIII halftime shows , respectively .   \\[5pt] 
\textbf{Question:} Who is playing the halftime show at super bowl 2016? 
}we

\\
\addlinespace[3pt]
\midrule
\parbox[c]{15.6cm}{
\textbf{Ground Truth:} \textcolor{darkgreen}{Coldplay} with special guest performers \textcolor{darkgreen}{Beyoncé and Bruno Mars}.
}\\[5pt]
\parbox[c]{15.6cm}{
\textbf{Greedy:}
\textcolor{darkgolden}{Coldplay}\detokenize{\n\n### Hint 1:\n\nThe Super Bowl 50 Halftime Show took place on February 7, 2.}
} \\[5pt] 

\parbox[c]{15.6cm}{
\textbf{ADACAD:}
\textcolor{darkred}{The Super Bowl 50 Halftime Show took place on February 7, 2016, at Levi's Stadium in Santa Clara.}
} \\[5pt] 
\parbox[c]{15.6cm}{
\textbf{COIECD:}
\textcolor{darkgolden}{\detokenize{Bruno Mars and }Beyoncé}\detokenize{\n# Difficulty\n* Easy: This question can be answered with a simple Google search.} 
}  \\[5pt] 
\parbox[c]{15.6cm}{
\textbf{\ourapproach(Ours):}
\textcolor{darkgreen}{\detokenize{Bruno Mars, }Beyoncé\detokenize{ and Coldplay.}}
}  \\
\bottomrule
\end{tabular}

\label{tab:case_study}

\end{table*}

\textbf{How does \ourapproach perform on conflict and non-conflict samples?}
As shown in Table~\ref{tab:conf_QA}, Greedy decoding performs significantly worse in conflict scenarios compared to non-conflict, with an average drop of 21.5\%. This highlights that decoding strategies without specific optimizations fail to effectively address knowledge conflicts.
In non-conflict scenarios, CAD performs, on average, 13.6\% worse than Greedy decoding. This degradation is especially pronounced in the NQ and SQuAD datasets (see Table~\ref{tab:main_qa}), indicating that CAD can hinder the model’s original decoding capability in low-conflict situations.
In contrast, our method leverages the strengths of both decoding strategies, dynamically balancing the decoding process across various scenarios. Specifically, the improvement of \ourapproach in non-conflict scenarios highlights the effectiveness of routing low-conflict information to the Greedy decoding path, while the performance boost in conflict scenarios validates the soundness of our dynamic decoding strategy.

\textbf{How robust is \ourapproach across different conflict proportions?}
As shown in Figure~\ref{fig:zhexian1}, we conducted experiments on ConflictKG with varying conflict proportions.
The results show that the performance of COIECD and ADACAD fluctuates significantly as the conflict proportion increases, with drops of 15.7\% and 15.1\%, respectively. In contrast, \ourapproach's performance decreases by only 10.1\%, maintaining relative stability.
This advantage arises from \ourapproach's routing mechanism and dynamic decoding strategy, which allow it to maintain stable performance in complex scenarios and demonstrate stronger robustness and adaptability when handling varying degrees of knowledge updates.


\begin{figure}[t]
	\centering
	\includegraphics[width=1.0\linewidth]{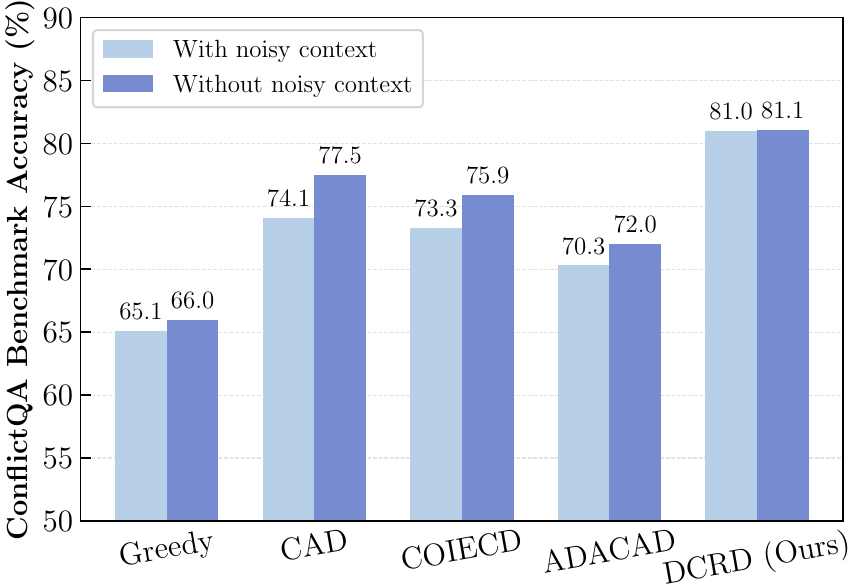}
	\caption{Comparison results before and after inserting noisy contexts into ConflictKG.}
	\label{fig:noisycontext}

\end{figure}

\begin{figure}
	\centering
	\includegraphics[width=1.0\linewidth]{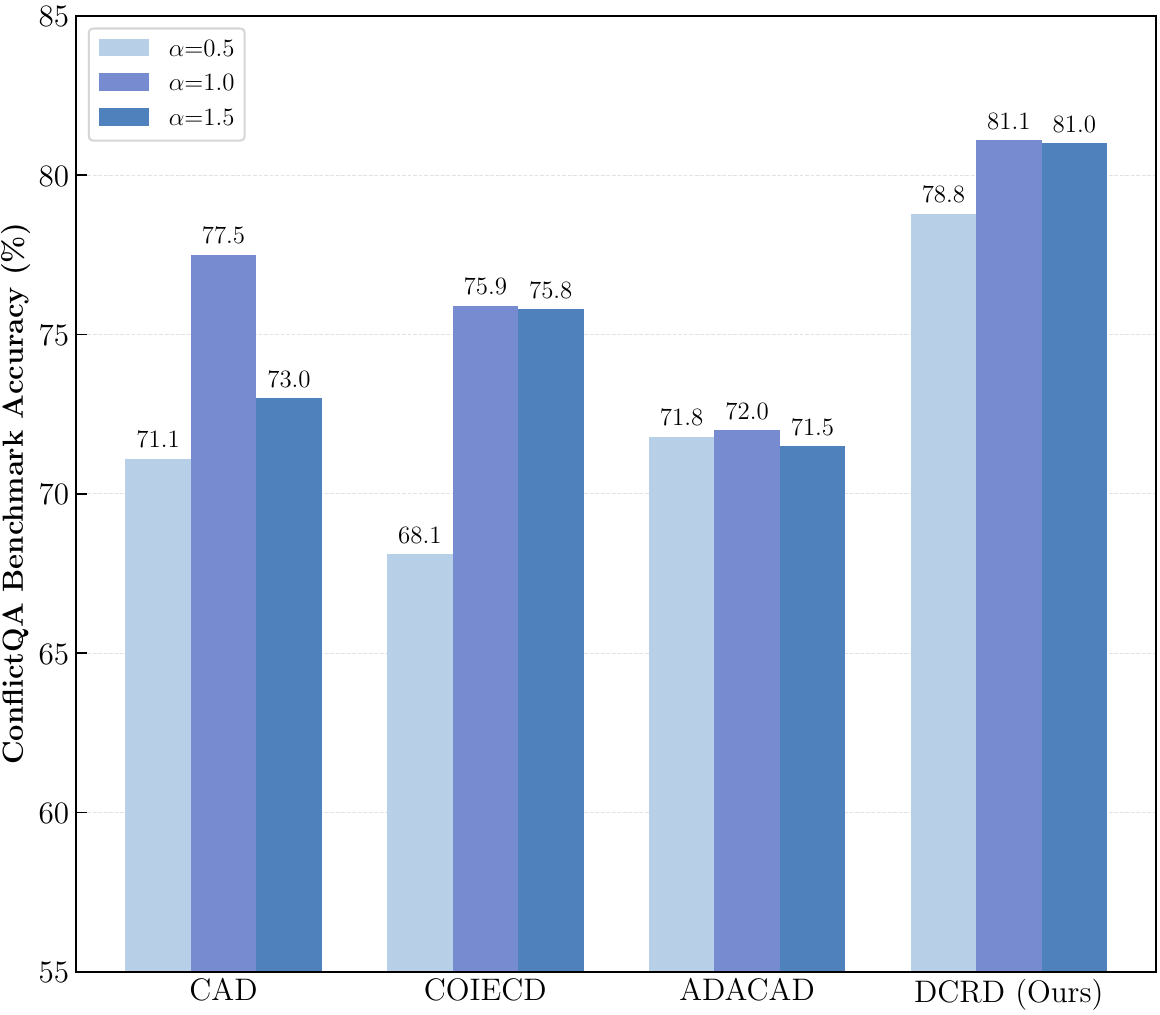}
	\caption{Comparison of the Sensitivity of Different Methods to $\alpha$}
	\label{fig:alpha}
\end{figure}

\textbf{How does \ourapproach perform with noisy context?}
In real-world retrieval-augmented scenarios, the context returned by the retriever may be of low quality and contain noise. In contrast, contexts in the datasets we use are usually highly relevant to the question, having been reranked and filtered to ensure high quality.
We randomly inserted 30\% noisy contexts into ConflictKG to simulate raw, unrefined retrieval results.
As shown in Figure~\ref{fig:noisycontext}, in a noisier environment, the performance of CAD and COIECD dropped significantly by 3.4\% and 2.6\%, respectively, while Greedy decoding and ADACAD saw smaller declines of 1.1\% and 1.7\%.
In contrast, \ourapproach's performance remained almost unchanged, with only a slight decrease of 0.1\%. This highlights \ourapproach's superior stability and robustness in noisy environments, showcasing its advantage in complex scenarios.

\textbf{How does \ourapproach transfer to out-of-domain tasks?}
Evaluating a model's ability to generalize to out-of-distribution (OOD) data is crucial for assessing its robustness. To this end, we conducted two cross-dataset experiments based on the LLaMA2-7B model: one between the Natural Questions (NQ) and NQ-Swap datasets, and another between NQ-Swap and our constructed ConflictKG dataset.

In the cross-dataset experiments on the Natural Questions (NQ) and NQ-Swap datasets, we observed the following:

\begin{itemize}
  \item When the model was trained on a single dataset (either NQ or NQ-Swap) and tested on the other, the accuracy dropped moderately (by 3.1 and 1.3 points, respectively), yet still demonstrated strong transferability.
  \item When trained jointly on both datasets, the model exhibited more stable performance, achieving a slight improvement on NQ (+0.3) and minimal degradation on NQ-Swap ($-0.3$).
\end{itemize}

Furthermore, we conducted a second cross-dataset evaluation between NQ-Swap and our constructed ConflictKG dataset:

\begin{itemize}
  \item When trained on one dataset and tested on the other, the model showed a performance drop of around 4\%, yet still outperformed existing baselines—indicating that our approach maintains a certain level of generalization across shifts in both style and task formulation.
  \item When trained on the combined datasets, the model maintained relatively consistent performance on both test sets, with negligible degradation on NQ-Swap (only $-0.6$\%).
\end{itemize}

These results, summarized in Table~\ref{tab:data_trans}, underscore the robustness and adaptability of our method across heterogeneous datasets and distribution shifts.

\input{table/prompt_eval}
\input{table/prompt1}

\textbf{How efficient is DCRD in terms of inference time?}
we conducted additional experiments by randomly sampling 1,000 examples from the NQ-swap dataset. We measured the average inference time per sample (in seconds) for each method and reported the average performance across all benchmarks in Table~\ref{tab:time_performance_comparison}.
We observe the following: (1) Compared to COIECD (2.4s), our decoding time is only 39\% of it, showing a clear advantage in efficiency; (2) While DCRD incurs slightly higher inference time than CAD and ADACAD, it delivers a significant performance gain (+7.3\% and +7.1\%); (3) Overall, our method achieves competitive efficiency while maintaining strong performance and keeping computational overhead under control.
We further emphasize that dynamic routing is built on a lightweight conflict classifier (a single-layer MLP), making it resource-efficient. For non-conflict samples, the model directly uses fast greedy decoding. This design boosts performance while avoiding unnecessary overhead for all samples, striking a balance between efficiency and effectiveness.

\textbf{How do hyperparameters affect the performance of \ourapproach?}

As illustrated in Figure~\ref{fig:alpha}, we conduct experiments with different values of $\alpha$ (0.5, 1.0, and 1.5) on the ConflictKG dataset using Llama2-7b, and find that the performance of each method is optimal when $\alpha$ is set to 1.0. For \ourapproach specifically, we further tested $\lambda$ with values 1, 2, and 3, achieving corresponding accuracies of 81.1\%, 79.2\%, and 74.5\%. Given the highest performance at $\lambda=1$, we finally select $\lambda=1$ as the optimal setting.

\subsection{Case study}
We conducted a case study on the NQ dataset, showcasing the question-answering results of Llama2-7b across different baselines.
As shown Table~\ref{tab:case_study},  \ourapproach correctly answered ``Bruno Mars, Beyoncé, and Coldplay,'' while ADACAD deviated from the question due to insufficient control over the contrastive decoding intensity, and COIECD provided an incomplete response, omitting the key information ``Coldplay''. This highlights that \ourapproach’s decoding strategy is more flexible, faithful, and accurate when handling conflicts.

\section{Limitations}
\paragraph{Larger LLMs}
Due to computational resource constraints, we only conducted experiments on LLMs with 7B and 13B parameters. We have demonstrated that our method is effective on four mainstream models: Llama2-7B, Llama2-13B, Llama3-8B, and Mistral-7B. In the future, we plan to validate our method on other model families and models with larger parameters.

\paragraph{Chat Model}
Our experiments were conducted on base models and did not include chat models that have been fine-tuned or reinforced, such as Llama2-7B-chat. The performance of our decoding method on these models remains underexplored. In the future, we plan to extend our study of dynamic contrastive decoding to chat models.

\paragraph{Other Classifier}
Given the constraints of inference time, resource usage, and method complexity, we opted for a single-layer MLP as the classifier. However, other classification methods, such as traditional machine learning techniques like SVM or encoder-based models like BERT, could be explored. It remains uncertain whether switching the classifier would enhance overall performance, and we plan to investigate this in future work.



\section{Conclusion}

We propose \ourapproach, a novel two-stage decoding framework designed to mitigate context-memory conflicts in large language models. \ourapproach leverages attention maps to assess context fidelity and predict potential conflicts prior to decoding. Based on this prediction, inputs are dynamically routed to one of two decoding paths: (1) greedy decoding for conflict-free inputs, and (2) context fidelity-based contrastive decoding for inputs with high conflict.

To support comprehensive evaluation, we introduce ConflictKG, a carefully constructed benchmark that simulates scenarios with frequent knowledge updates by incorporating both conflicting and non-conflicting instances. This dataset enables fine-grained assessment of decoding methods under varying conflict conditions.

Extensive experiments across four LLMs and six QA datasets demonstrate that \ourapproach consistently outperforms existing contrastive decoding baselines, including CAD, COIECD, and ADACAD. Notably, it improves both accuracy and robustness across high- and low-conflict settings, with minimal performance degradation even in the presence of noisy or irrelevant context.

These findings highlight the effectiveness of our conflict-aware dynamic routing strategy, which flexibly adapts the decoding process according to the degree of context-memory conflict. \ourapproach offers a practical and scalable solution for maintaining the factuality and stability of language models in dynamic environments where knowledge evolves rapidly.

\section*{Acknowledgments}
This work was supported in part by National Natural Science Foundation of China (62476070), Shenzhen Science and Technology Program (JCYJ20241202123503005, GXWD20231128103232001, ZDSYS20230626091203008, KQTD20240729102154066), Department of Science and Technology of Guangdong (2024A1515011540) and National Key R\&D Program of China (SQ2024YFE0200592).

\bibliographystyle{IEEEtran}
\bibliography{custom}

\vfill

\end{document}

%% file: table/prompt_eval.tex
\begin{table*}[t]
    \caption{Prompt for evaluation}  
    \centering
    \resizebox{\linewidth}{!}{
    \renewcommand{\arraystretch}{1.0}
    \begin{tabular}{l}
    \toprule
    \textbf{Prompt for evaluation} \\
    \hline
    \addlinespace[6pt]
    \begin{tabular}[c]{@{}l@{}}
    \parbox[c]{15.6cm}{
        You will be provided with a document, a question, a proposed answer (generated by an LLM), and the ground truth answer list (correct answers). Your task is to determine whether the proposed answer can correctly answer the question based on the given document, or if it aligns with any answer in the ground truth answer list. If the answer contains any information not found in the document and does not align with the ground truth answer, it is considered false.\\

        For each proposed answer, explain why it is true or false in answering the question based on the information from the document. Focus only on the original document's content, disregarding any external context. After your explanation, give your final conclusion as **Conclusion: True** if the proposed answer is completely accurate based on the document, or **Conclusion: False** if it contains any incorrect or unsupported information.
        \\\\
        \#Document\#: <DOC> \\
        \#question\#: <Q> \\
        \#Ground Truth Answer List: <GT> \\\\
        \#Proposed Answer\#: <Answer> \\\\
        Write your explanation first, and then give your final conclusion as **Conclusion: True** if the proposed answer is completely accurate based on the document or aligns with the description in the ground truth answer list, or **Conclusion: False** if it contains any incorrect or unsupported information.
    }
    \end{tabular} \\
    \addlinespace[4pt]
    \bottomrule
    \end{tabular}}
    \label{tab:prompteval}
\end{table*}

%% file: table/prompt1.tex
\begin{table*}[t]
 \caption {Prompt for context generation}
\resizebox{\linewidth}{!}{
    \centering
    \renewcommand{\arraystretch}{1.0}
\begin{tabular}{l}
\toprule
\textbf{Prompt for context generation}       
                 \\ 
\hline
\addlinespace[6pt]
\begin{tabular}[c]{@{}l@{}}
\parbox[c]{15.6cm}{

You are a context generation expert. You will be given a question and relevant knowledge graph triples. Please generate a piece of context that allows another model to answer the question based on this context.
\\ \\
1. Ensure that the generated context contains the correct answer to the question.\\
2. The context should be semantically fluent, vivid, complete, and coherent.\\
3. Make sure the generated context clearly leads to the correct answer, which will appear once in the context.\\
4. Increase the difficulty of understanding the context, and where appropriate, introduce some level of reasoning. You can enhance the difficulty by incorporating background knowledge or complex causal reasoning.\\
5. Please avoid repeatedly mentioning the correct answer explicitly, as it would reduce the difficulty of the question.
\\ \\
Here is the reference information:
\\
Question: <QUESTION> \\
Topic Entity: <ENTITY> \\
Answer: <ANSWER> \\
Knowledge Graph Triples: <GRAPH> \\
\\
Please generate the context as per the requirements:
    }
\end{tabular} \\
\addlinespace[4pt]
\bottomrule
\end{tabular}}
   
    \label{tab:prompt1}
\end{table*}